\documentclass{article}
\usepackage[utf8]{inputenc}
\usepackage{float}
\usepackage{changepage}
\pdfoutput=1

\makeatletter
\newcommand*{\rom}[1]{\expandafter\@slowromancap\romannumeral #1@}
\makeatother

\newenvironment{tightcenter}{%
  \setlength\topsep{0pt}
  \setlength\parskip{0pt}
  \begin{center}
}{%
  \end{center}
}

\title{The Computational Principles of Learning Ability}
\author{Wu Hao\\
email:wuhao29@gmail.com}
\date{September 2015}

\usepackage[numbers]{natbib}
\usepackage{graphicx}

\begin{document}

\maketitle
\begin{abstract}
It has been quite a long time since AI researchers in the field of computer science stop talking about simulating  human intelligence or trying to explain how brain works. Recently, represented by deep learning techniques, the field of machine learning is experiencing unprecedented prosperity and some applications with near human-level performance bring researchers confidence to imply that their approaches are the promising candidate for understanding the mechanism of human brain\cite{le2013building}\cite{cadieu2014deep}. However apart from several ancient philological criteria and some imaginary black box tests (Turing test, Chinese room) there is no computational level explanation, definition or criteria about intelligence or any of its components. Base on the common sense that learning ability is one critical component of intelligence and inspect from the viewpoint of mapping relations, this paper presents two laws which explains what is the ``learning ability" as we familiar with and under what conditions a mapping relation can be acknowledged as ``Learning Model".
\end{abstract}

\section{Introduction}
\begin{figure}[H]
    \centering
    \includegraphics[width=120mm,scale = 0.5]{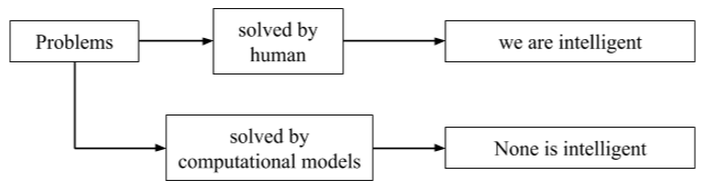}
    \caption{Paradox of AI effect}
    \label{paradox}
\end{figure}

Except for many philosophical discriptions and the famous black box test “Turing test”, there is no clear definition of intelligence and it is well accepted that the ability of “thinking” is difficult to define\cite{turing1950computing}. On the other hand, for machine learning researchers and artificial intelligence experts, once a problem is solved, the solution as a computational model, seems to have nothing to do with intelligence, it seems like only a problem to be solved is related to the understanding of intelligences\cite{mccorduck2004machines}. Therefore, researchers are trapped in a paradox as shown in figure \ref{paradox}.

\begin{figure}[H]
    \centering
    \includegraphics[width=120mm,scale = 0.5]{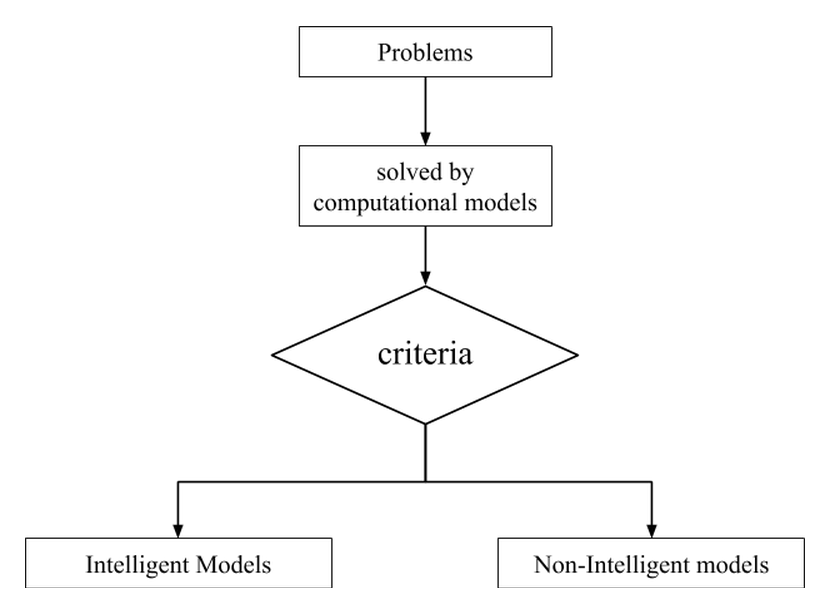}
    \caption{White box criteria for Intelligent Models}
    \label{4intelligent}
\end{figure}

\begin{figure}[H]
    \centering
    \includegraphics[width=120mm,scale = 0.5]{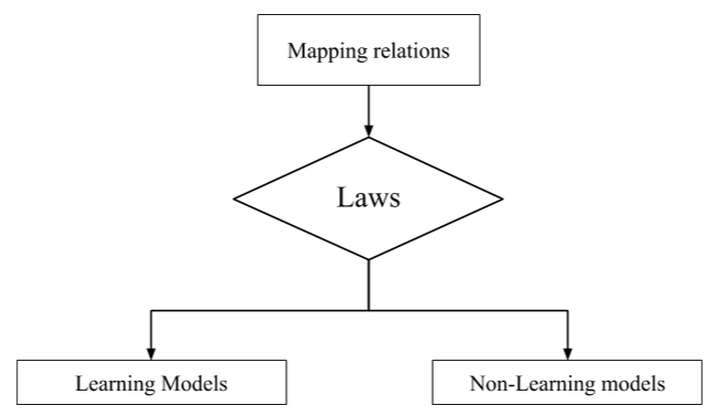}
    \caption{White box criteria for Learning model}
    \label{4learning}
\end{figure}

\noindent
Since all possible automated solutions implemented by computer systems are basically different computational models\cite{berlinski2000advent}, it seems like there will be no computational model which could be acknowledged as possessing true intelligence forever. One possible solution of breaking this paradox is to find one or a set of criteria which can be used for white box testing of all computational models, as shown in figure \ref{4intelligent}.
\noindent
Instead of giving criteria for intelligences, based on the understanding that the learning ability is a critical component of intelligence, this paper proposes two laws for a computational model to be a learning model. With the help of these two laws, computational models can be classified as ``Learning Model" and ``Non-Learning Model" (figure \ref{4learning}), these two laws also provide a computational explanation about what the ``Learning Ability"  is.

Based on the viewpoint that has been missed by previous researchers, this paper focuses on discussing the behaviour of a model, or more specifically the behaviour of individual configurations  $\theta \in \Theta$ of a given dynamic system $M(x,\Theta)$ ($\Theta$ is the parameter set and $\theta$ is any of its element, the detailed discussion of dynamic systems is in the explanatory part of section three).  The ``Learning Ability" discussed in this paper is a behaviour of a mapping relation, so when being used in the term ``Learning Model" and ``Machine Learning" (Learning Theory), the word ``Learning" carries different meanings respectively.

It is a common sense that without learning ability a creature cannot be called as an intelligent creature. And what being learned is nothing but information, the next section gives six definitions about relations between information and mapping relation. And only by expressing some common philosophical understanding about information formally is further inference possible, these six definitions play key roles in understanding these two laws and corresponding corollaries. Section three introduces two laws that explain the fundamental mechanism of our intuitive feeling of learning ability and three corresponding corollaries that proves the existence of a common learning model which satisfies these two laws. These two laws and corollaries also illustrate several core properties of this common learning model. 

\section{Definitions}

\subsection{Definition \rom{1}}

\begin{tightcenter}
\textbf{
        \textit{
For a given mapping relation $M:D\rightarrow O$, an element of the range  (
 $o \in O$)  and 
 each element in its corresponding domain ($\{e \mid e \in D, M(e)=o \}$)
are global and local information respectively.
                }
        }
\end{tightcenter}

\bigskip

When we sit on our chair and look outside the window, we could see birds, butterflies, clouds, and trees. When we take a deep breath, we could taste the sweet smell of the freshly cut grass. Although we have no direct access to the world other than through our sensors \cite{luft2011subjectivity}, we can always rely on different kinds of apparatus to discover the world, in fact all apparatus can be regarded as extension of our biological sensors. But what if something cannot be detected by all means? Is it necessary to insist on its existence? The question has been answered perfectly by Carl Sagan's famous story \textit{“The Dragon In My Garage”}\cite{sagan2011demon}. And the following notion has been well known and accepted for decades.

\begin{tightcenter}
\textbf{
        \textit{
        There is no observation independent reality.
                }
        }
\end{tightcenter}

\noindent This statement illustrates the relation of two kind of information:

\begin{enumerate}
\item Observation $\Rightarrow$ a set of appearances
\item Reality $\Rightarrow$ being defined by a set of appearances
\end{enumerate}

\noindent Any apparatus being used to detect the world can be regarded as creating a mapping relation from appearances (subset of apparatus' domain) to the reality (elements in the range) and biological systems can also be regarded as one kind of apparatus which includes us.

\bigskip

\noindent The term ``reality" means the existence of certain concept which could be an object or an abstract concept such as the existence of gravity or electromagnetic wave. Therefore the term ``reality" will be replaced by ``concept" in the following sections.

\subsection{Definition \rom{2}}
\begin{tightcenter}
\textbf{
        \textit{For a given mapping relation $M:D\rightarrow O$, one subset of the domain and each element of the corresponding subset of the range are local representation and global representations which define the same concept respectively.
                }
        }
\end{tightcenter}

\begin{figure}[H]
    \centering
    \includegraphics[width=80mm,scale = 0.5]{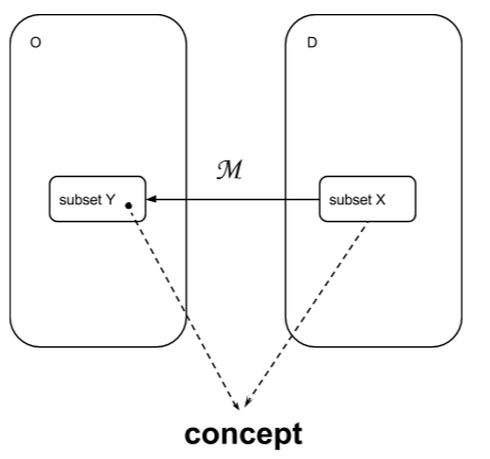}
    \caption{Local and Global representation}
    \label{def2}
\end{figure}

\noindent A set of appearances are detected by an apparatus M (the domain of a mapping relation M). All these appearances indicate the existence of a concept, so the local representation of this concept is defined as the subset X and the global representation of this concept is defined as each element of the subset Y. Therefore all local information about a concept is its only local representation and the global information of the concept is equivalent to its global representation.

\bigskip
\noindent In summary, definition \textit{\rom{1}} and definition \textit{\rom{2}} indicate that:

\begin{enumerate}
\item These four expressions are equivalent:\\
A concept$\equiv$A set of appearances$\equiv$A set of local information (definition one)$\equiv$A local representation (definition two)

\item These three expressions are equivalent:
A concept$\equiv$A piece of global information$\equiv$A global representation
\end{enumerate}

\subsection{Definition \rom{3}}
\begin{tightcenter}
\textbf{
        \textit{
        \begin{flushleft}
        For a given mapping relation $M:D\rightarrow O$ and subset $X \subset O$, we have $Y = \{y \mid y=M(x), x\in X\}$, so that:\\
        \end{flushleft}
        \begin{enumerate}
        \item $\forall y \in Y $subject to a constraint $c_\alpha$ which is defined as a pair  $<Y,R_i>$, $ R_i$ is the i-ary relation;
        \item $\forall x \in X $subject to a constraint $c_\beta$ which is defined as a pair  $<X,R_j>$, $ R_j$ is the j-ary relation;
        \item if $R_i$ and $R_j$ are different, then each element of Y is a invariant representation of X.
        \end{enumerate}
                }
        }
\end{tightcenter}

\noindent
Together definition \textit{\rom{1}} and \textit{\rom{2}} indicate that there could be more than one global representation for one concept, therefore in order to guarantee the certainty of information, all global representations are supposed to be subject to at least one constraint, or in other words, it is necessary to find the constraint that all global representations are subject to so that these global information can be recognised or be harvested. If all global representations of a concept and all appearances of this concept are only subject to different constraints, then global representation of this concept is its invariant representation.

\subsection{Definition \rom{4}}
\begin{tightcenter}
\textbf{
        \textit{
        For a given mapping relation $M:D\rightarrow O$, it defines the type of global information as $M$. 
                }
        }
\end{tightcenter}
\bigskip

\noindent 
As the notion being introduced in definition \textit{\rom{1}}:

\begin{tightcenter}
There is no observation independent reality
\end{tightcenter}

\noindent
The existence of a certain concept depends on whether it is observable. Furthermore the nature of the concept depends on the method of observation.

\bigskip
\noindent
Definition \textit{\rom{4}} guarantees that it is the appearances and the way these appearances are being processed that decide not only the existence but also the nature of the concept, because the appearances (local information) form a subset of the domain, and the way these local information are being processed (mapping relation) gives the global representation (global information) of a system(the concept). This definition is also a generalisation of our daily life experiences.
 
\bigskip
\noindent
Visual detection enables us to tell different kinds of trees. And it is hard to tell the differences of these realities by only touching them. Because with the change of the domain, the appearances of these realities do not carry enough information for telling differences within each kind, but we can still tell the differences between plant and animal by touching them.

\bigskip
\noindent
Even when facing the same domain, different mapping relations will give different types of information. One typical example is the camera where the domain is provided by the CCD array, and different functions of the camera will give different types of information, such as the focusing information being used to adjust the lens and the information being recorded as photos.

\subsection{Definition \rom{5}}
\begin{tightcenter}
\textbf{
        \textit{
        For two given mapping relations: $M_l:S_1\rightarrow S_2$ and $M_h:S_3 \rightarrow S_4$, if $S_1 \cup S_2 \subset S_3$ then $  \forall s \in S_2$ is homologous global information with respect to $M_h$.
                }
        }
\end{tightcenter}
\begin{figure}[H]
    \centering
    \includegraphics[width=50mm,scale = 0.5]{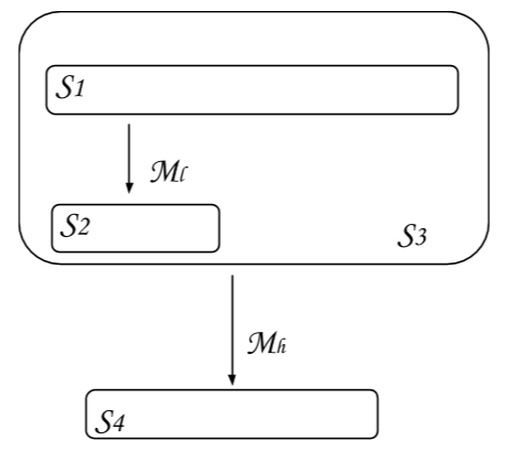}
    \caption{Homologous Global Information}
    \label{def5}
\end{figure}

\subsection{Definition \rom{6}}
\begin{tightcenter}
\textbf{
        \textit{ For two given mapping relations: $M_l:S_1\rightarrow S_2$ and $M_h:S_3 \rightarrow S_4$, if $S_2 \subset S_3$ then $ \forall s \in S_4$ is first order global information with respect to $M_l$.
                }
        }
\end{tightcenter}
\begin{figure}[H]
    \centering
    \includegraphics[width=50mm,scale = 0.5]{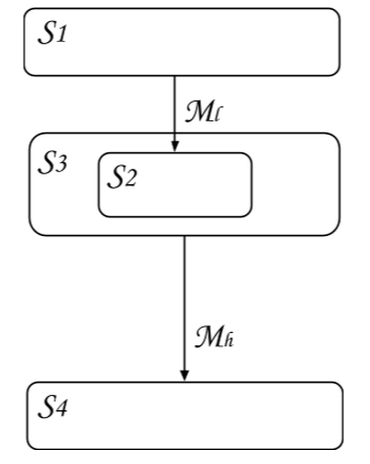}
    \caption{First Order Global Information}
    \label{def6}
\end{figure}
Apparently, all elements in $S_4$ is beyond the field of vision of mapping relation $M_l$, actually if $M_h$ is one to one mapping relation between $S_2 = S_3$ and $S_4$, it is also true that all elements  in $S_2$ is beyond the field of vision of mapping relation $M_l$. Therefore for a given mapping relation $M:D \rightarrow O$, all elements in its range is first order global information with respect to $M$ itself. And together with definition \textit{\rom{2}} we know that these three concepts are equivalent: a global information, a global representation, a first order global information.

\bigskip
\noindent
Usually a mapping relation is the composition of several mapping relations, it could be combination of different approaches of processing information from the domain. Definitions \textit{\rom{5}} and \textit{\rom{6}} describe the relation between different level of process.

\subsection{Explanatory Comment}
This section explains in what sense these definitions shall be required to be understood. Our daily experience are further processed products of basic information and two trivial facts of information are usually being ignored, yet they are the most notable, because these two facts lead to a unique viewpoint which will enable us to understand the relationship between the appearance and the reality, and the role of constraint. The understanding of these concepts is critical for establishing the theoretical framework which can be used to explain what learning is, analyse the learning ability of a given model and construct learning models with different levels of ability.

\begin{enumerate}
\item The concept of global and local are comparative.

More specifically, it means the existence of a concept depends on its appearances, and this concept itself could also be one of many appearances which define a higher level concept. This can be explained by a simple mental experiment. Imagine you were sleeping on the backseat of a minivan, the shaking caused by the speed bump wakes you up. You do not know how long you have slept and you watch the scenery passing outside the window with your sleepy eyes. Suddenly you realised that you are approaching ``Some Place". What is included in the passing landscape depends on their appearance, it could be a cottage, a church or a supermarket. And all these rapidly passing views are appearances which enable you to recognise that you are near this “some place”, and this ``some place" is a higher level concept. Therefore, it is necessary to give definitions which can be used to describe this hierarchical relationship of different information.

\item Certainty plays a critical role in representing information, in fact without certainty there will be no information. Even when people are measuring the uncertainty of a system ( entropy), information of the number of possible states and the probability of being in each state are still necessary and must be expressed in a certain way:

\begin{figure}[H]
    \centering
    \includegraphics[width=70mm,scale = 0.5]{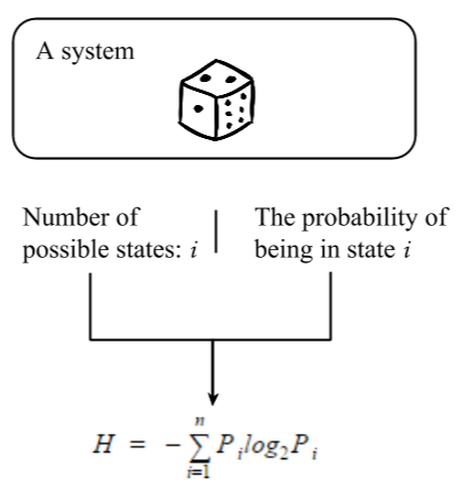}
    \caption{Information needed for calculating entropy}
    \label{entropy}
\end{figure}

\noindent
If the precise value of $i$ and $P_i$ cannot be obtained, we still need a certain number to represent the information of a certain within bounds If there is no information about $i$ and $P_i$ at all, there is still a piece of notation which represent the definition of entropy; and if there was nothing in figure  \ref{entropy} then there will be no information at all. Therefore it is reasonable to say that one piece of information must be carried by at least one certain state and one certain state can be used to represent at least one piece of information. 

However, certainty itself only guarantees the existence of information, whether the information can be recognised relies on another critical factor: constraint. This relation between certainty and constraint is usually being ignored because most of the time in our daily life, the acquisition and recognition of information occur one after another immediately and the harvested information directly affects all of our sensation and perception. Occasionally people could have the chance to realise the existence of this relation, such as when we are hearing people speaking another language for the first time. Because of being subject to different constraint, for example, it is hard for most readers of this article to understand Chinese. And we have not known exactly what the constraint is for any language, although we rely on at least one of these constraints everyday. 

\end{enumerate}
Relying on variance hypothesises of different constraints, researchers try to mimic unknown constraints of different languages or the world we are seeing, the development of this field brings various NLP and classification applications, some with great performance. Almost all researchers in this field claim that they are interested in knowing how the brain might work and imply their applications are promising candidates that could lead us on the course to the answer. Therefore, one question to which we are all interested to know the answer is ``Are these applications qualified to be prefixed with intelligent?". The following is an imaginary experiment which illustrates one of many possible key factors of this question:
\bigskip
\indent 
\begin{tightcenter}
        \begin{flushleft}
        \textbf{Suppose people were able to create today's state-of-the-art on-line translator (such as Skye on-line translator) in the beginning of this millennium and keep it running. Could this system be able to translate ``selfies" into Chinese now? 
                }
        \end{flushleft}
\end{tightcenter}
\bigskip
\noindent
If the system could, does this mean that the system can be labeled as ``intelligent" indisputably? After all, at the time the system was created, the words `selfies' had not been invented and the system somehow managed to learn the ``meaning" of this word, therefore it is possible that the system could be as smart as human. Now the focus of the argument about ``intelligence" switches to exactly how this information was learned by the system. What if the system was linked to a dictionary which was periodically updated by human? what if the system asks end users to report mistakes and automate the correction process? Actually, no matter what strategies are being used, it must be some kind of computation, but exactly what kind of computation would match our understanding of learning? The answer to this question is: ``There is no criterion of learning ability for computational models yet" or in other words, we have no formal definition of learning for computational model. Therefore the judgement of learning ability for computational model cannot be made, this is also part of the reason why, as Rodney Brooks complained\cite{mcginity2004s}:

\begin{tightcenter}
        \textit{``Every time we figure out a piece of it, it stops being magical; we
say, ’Oh, that’s just a computation"
                }
\end{tightcenter}

Usually human do not start talking before one-year-old, we cannot recognise individuals of different species or even different races unless we have seen enough, we train dogs to help us because we know dogs are smart animal, meanwhile we also like funny videos which record dog confused by ``invisible doors". It is highly unlikely that the effort of understanding what is intelligence would bring us C3PO directly, the lack of a criterion for learning ability of computation model would make us ignore some computational models which could be a building block of higher level learning ability or intelligence, only because their current performance is not as good as some other computation model which might have no such potential. 

And only with the help of such criteria could we carry out research to help us understand the nature of learning, analyse the level of learning ability quantitatively, construct computational model with different levels of learning ability, and eventually provide people the research materials for studying what is intelligence, and the intelligence of us. Based on definitions about information given in this section, next section will discuss two laws for a mapping relation to be a learning model and the corresponding corollaries.

\section{Axiom and The Laws of Learning}
\subsection{Axiom}
\begin{tightcenter}
\textbf{
        \textit{
        Without mapping relation there will be no acquisition of any information.
                }
        }
\end{tightcenter}
This axiom comes from interpreting the statement at the beginning of section two:
\begin{tightcenter}
    \textit{
        There is no observation independent reality.
            }
\end{tightcenter}
\bigskip
\noindent
In other words, this axiom indicates that mapping relation is necessary for defining concepts.
And if a mapping relation $M_L: D \rightarrow O$ is said to be a learning model which is able to learn from its domain D, it must follow these two laws.
\subsection{Law \rom{1}}

\begin{tightcenter}
\textbf{
        \textit{
        $\forall X_S \subset D, Y_S = M_L(X_S), \exists X_N \subset D \setminus X_S : Y_N = M_L(X_N), Y_N \cap Y_S = \emptyset$
                }
        }
\end{tightcenter}
        
        \begin{flushleft}
        For any subset $X_S$ of the domain $D$, and the corresponding subset of the range is $Y_S$, there exist a set $X_N$ which is a subset of the complement set of $X_S$, so that we have a subset $Y_N$ which is disjoint with $Y_S$.
        \end{flushleft}
    
\bigskip
\noindent
This law would be best understood by assuming that there is a mapping relation $M_{non}$ which does not obey law one. Then there will be four\footnote{Scenario two and three are equivalent.} possible scenarios as follows:

\begin{figure}[H]
    \centering
    \includegraphics[width=100mm,]{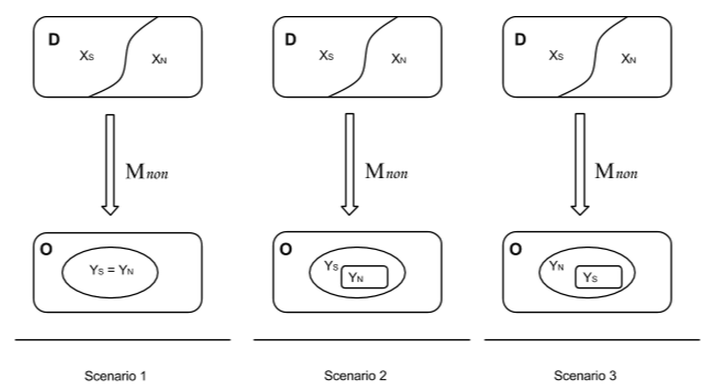}
    \caption{Interpretation of Law One: Scenarios 1-3}
    \label{law1_1}
\end{figure}

\noindent
In the three scenarios in figure \ref{law1_1}, $X_S$ and $X_N$ contain different local information, however the mapping relation $M_{non}$ will map them to the same set of global representations, therefore as a detector $M_{non}$ fails to detect different concepts in the domain. In other words, mapping relations which do not obey law one are basically information black holes which could allow the possible information of the existence of many concepts to devolve into the same state \cite[p.43]{anderson2015cosmic}.

\begin{figure}[H]
    \centering
    \includegraphics[width=60mm,scale = 0.5]{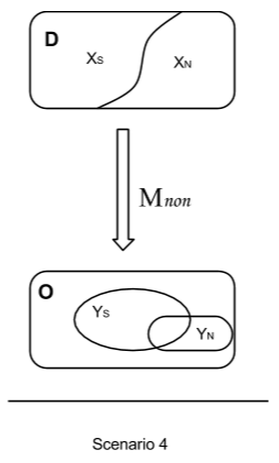}
    \caption{Interpretation of Law One: Scenarios 4}
    \label{law1_2}
\end{figure}

\noindent
In the scenario in figure \ref{law1_2}, for guaranteeing the certainty property of information, elements of $Y_S$ and $Y_N$ will subject to the same constraint\footnote{Trying to avoid using same constraint by labelling elements as described in above four scenarios will violate law \textit{\rom{2}}, and at this stage this violation seems inevitable. Formal mathematical discussion about this issue is part of the future works.}, therefore the possible existences of many concepts still devolve into the same state as in the previous three scenarios. 

\bigskip
\noindent
In summary, law \textit{\rom{1}} guarantees that the possible existence of a new concept will not be missed.

\subsection{Law \rom{2}}
\begin{tightcenter}
\textbf{
        \textit{The training process\footnote{The notation $\Lambda$ represents the parameter space of mapping relation $M_L$.} of $M_L(\alpha \subset \Lambda): D \rightarrow O$ should not depend on any of $M_L$'s first order global information $M_H$ which follows law \textit{\rom{1}} and law \textit{\rom{2}} as well.
                }
        }
\end{tightcenter}
\noindent
The intuitive explanation of law \textit{\rom{2}} is: a learning model should be able to define the existence of new concept all by itself. This explanation ends all similar arguments like ``Should a NLP system linked with a dictionary be label as 'Intelligence'?". Furthermore, together with law one, origins of information (mapping relations) being used for a composite mapping relation can be classified into two categories, learning model and non-learning model(memory system), so that further analysis could be possible. Actually whenever the construction process of a mapping relation is inferred by some given knowledge about the desirable learning results, it will almost always limit the learning ability. Topics related to law two are core problems which need further study.\footnote{ In this paper, this law is the only discussion which involves dynamicity of a mapping relation or usually being mentioned as ``Learning Process".}
\subsection{Corollary \rom{1}}
\begin{tightcenter}
\textbf{
        \textit{Given mapping relation $M_L: D \rightarrow O (O \subset R^n, D \subset R^m)$, if $M_L$ follows law one, then there exist a family of functions $H: R^n \rightarrow R$ which can be used to harvest information being contained in the range of $M_L$.}
        }
\end{tightcenter}

\subsubsection{Lemma One}
\begin{tightcenter}
\textbf{
        \textit{Given $ D \subset R^n$: $ \forall X_j \subset D$ and $\forall r \in R$, $\exists \Phi(x \mid x \in X_j) = r$.
                }
        }
\end{tightcenter}

\noindent
For any subset $X_j$ of a n-dimensional real number domain $D$ and any given real number $r$, there exist an equality constraint $\Phi (x)$, so that  $\Phi(x \mid x \in X_j) = r$.

\bigskip
\noindent \textit{Proof:}
\begin{adjustwidth}{2cm}{}
For a given subset $X_j \subset D$, there exists an equation\footnote{Solution of this equation could not be an element of $X_j$, this equation ``$G(t)$" is just an example which proofs the existence of equality constraint $\Phi$.}:
\begin{equation}
G(t) = (t-x_1) \bullet (t-x_2) \bullet ...\bullet(t-x_i)
\end{equation}
so that for any $x \in X_j$, we have $G(x) = 0$.
\\
And for a given real number $r$, there exists:
\begin{equation}
\Phi(t) = G(t) + r
\end{equation}
so that for any $x \in X_j$, we have $\Phi(x) = r$.
\end{adjustwidth}

\bigskip
\noindent
Lemma one shows that for a given subset $X_j$ of a n-dimensional real number space, there exists a family of equality constraints: $\Phi_{Xj}(t,r \mid X_j)$, so that $\forall x \in X_j$ and $ \forall r \in R$ we have $\Phi^r_{Xj}(x \mid r, X_j) = r$.

\begin{figure}[H]
    \centering
    \includegraphics[width=130mm]{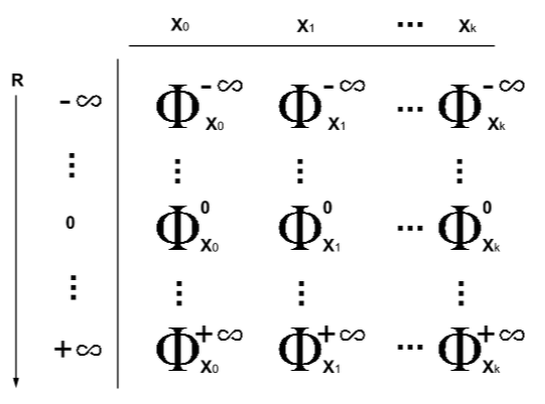}
    \caption{Constraint families defined by powerset of $D$}
    \label{lemma1_2}
\end{figure}

\bigskip
\noindent 
The constraint family $\Phi_{Xj}(t,r \mid X_j)$ can also be expressed as a column that includes infinitely many constraints:
\begin{figure}[H]
    \centering
    \includegraphics[width=60mm,scale = 0.5]{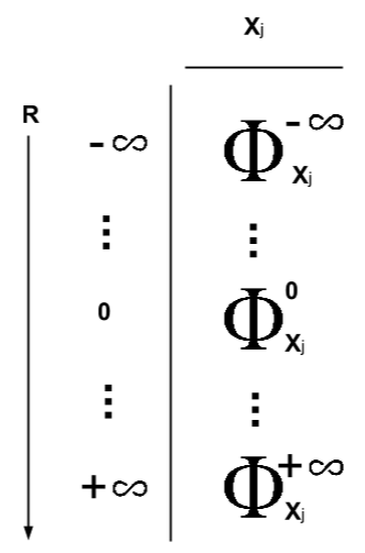}
    \caption{A family of constraints defined by $X_j$}
    \label{lemma1_1}
\end{figure}
\bigskip
\noindent 
Further more, constraint family $\Phi(t,r,X_k)$ is defined by the power set of the domain $D$ ($\forall X_k \in \mathcal{P}(D)$), each member of this family is $\Phi^r_{X_k}(x) = r$ and the constraint family $\Phi(t,r,X_k)$ can also be expressed wit a matrix as shown in figure \ref{lemma1_2}.

\bigskip
\noindent
\textit{Proof of Corollary One}
\\
\begin{adjustwidth}{2cm}{}
Known that mapping relation $M_L: D \rightarrow O (O \subset R^n, D \subset R^m)$ satisfies law \textit{\rom{1}}, so that for possible learning result $Y_1$ there exists an equality constraint $\Phi^{r1}_{Y_1}$, and for a new learning result $Y_2$, there exists infinity many equality constraints $\Phi^{r2 \neq r1}_{Y_2}$, so for a new learning result $Y_i$, there always exist an equality constraint $ \Phi^{ri}_{Y_i}$ which is defined by $Y_i$ and a real number $r_i$ ($r_i \neq r_1$, $r_i \neq r_2$). Then the constraint that all possible learning results follow can be expressed as:
\begin{figure}[H]
    \centering
    \includegraphics[width=90mm]{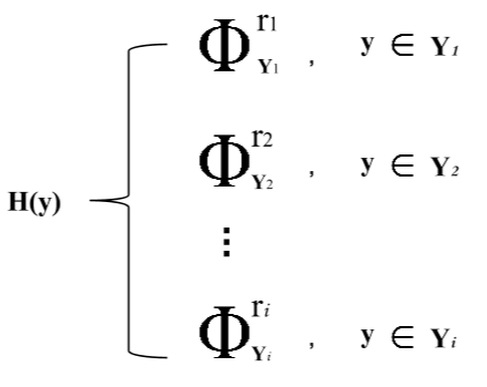}
    \caption{Information Harvesting Function $H$}
    \label{corollary1}
\end{figure}

\noindent
Because $Y_1$,$Y_2$ to $Y_i$ are disjoint sets(law \textit{\rom{1}}) and the corresponding formula $\Phi^{ri}_{Y_I}(y)$ equals a unique real number, therefore $H(y)$ is by definition a function. And there could be infinite many possible $H(y)$. Thus $H$ is the harvesting function which can harvest information from a mapping relation which follows law \textit{\rom{1}}.

\end{adjustwidth}

\subsection{Corollary \rom{2}}
\begin{tightcenter}
\textbf{
       The composition of mapping relations $M_L$ which follows law \textit{\rom{1}} and its corresponding harvesting function $H$ still follows law \textit{\rom{1}}. 
        }
\end{tightcenter}

Proof:
\begin{adjustwidth}{2cm}{}
Denote $F_l=H \circ M_L$.\\
Because $M_L$ follows law one then:\\
$\forall X_S \subset D, \exists X_N \subset D \setminus X_S$\\ so that $Y_S=M_L(X_S), Y_N = M_L(X_N)$ and $Y_N \cap Y_S = \phi$\\
Then according to the definition of (H):\\
$H(y\mid y \in Y_S) \neq H(y \mid y \in Y_N)$.\\
Therefore we know $\forall X_S \subset D, \exists X_N \subset D \setminus X_S$:\\
$F_L(X_S) \neq F_L(X_N)$
\end{adjustwidth}

\bigskip
\noindent
Corollary two indicates it is possible to represent different concepts of information type $M_L$ by using different real number.

\subsection{Corollary \rom{3}}

\begin{tightcenter}
\textbf{
        \textit{ 
The composition of mapping relations $M_L$ and any of its harvesting function $H$ is equivalent to a common constraint $V_C$ and a mapping relation "L" which also satisfies law \textit{\rom{1}}.
                }
        }
\end{tightcenter}

\begin{figure}[H]
    \centering
    \includegraphics[width=140mm]{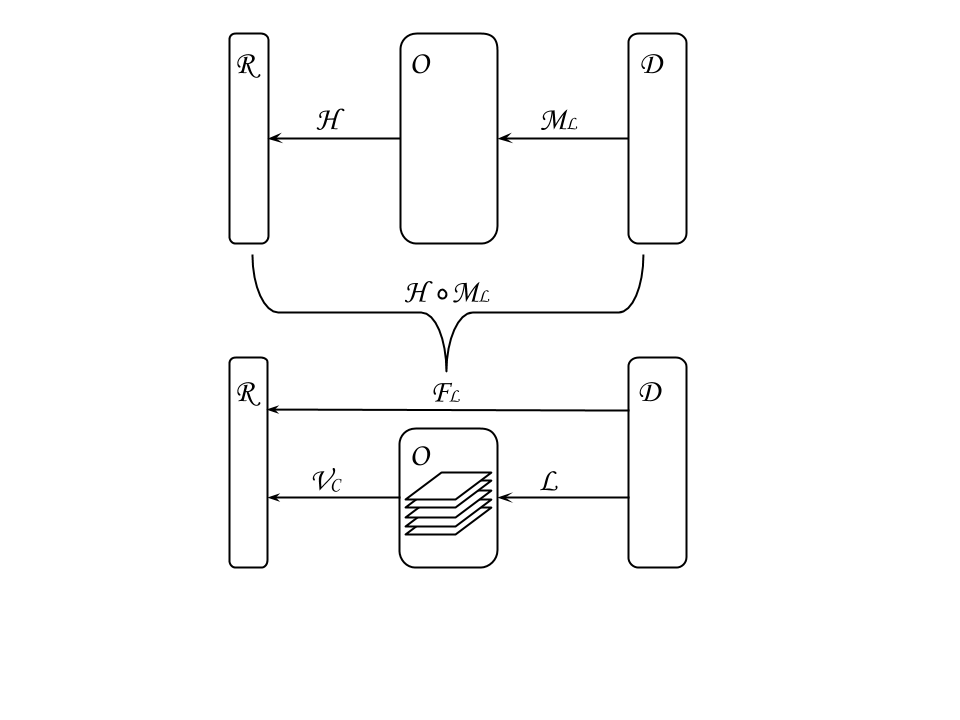}
    \caption{Common Constraint}
    \label{corollary3}
\end{figure}

Proof:
\begin{adjustwidth}{2cm}{}
Real numbers can be represent by parallel hyperplanes defined by a vector set $V_C$. And because of corollary \textit{\rom{2}} we know:\\
$\forall X_S \subset D, \exists X_N \subset D \setminus X_S$: $F_L(X_S) \neq F_L(X_N)$ (corollary two)\\
Therefore $L(X_S) \cap L(X_N) = \phi$ (satisfies law \textit{\rom{1}})
\end{adjustwidth}

\bigskip
\noindent
This corollary carries double meaning:
\begin{enumerate}
\item Linear separability is the common constraint for all mapping relations that satisfy law one could convert to.
\item Without the harvesting function, the learning result of mapping relation $M_L$ cannot be recognised, so the mapping relation $F_L$ (as shown in figure \ref{corollary3}) is supposed to be the mapping relation which could provide useful information eventually. However, in spite of the existence of infinitely many possible $H$, it is almost impossible to locate a suitable harvesting function without violating law two. Corollary three shows that $V_C$ is independent of any first order global information of $F_L$ (prior knowledge about the undetected concepts)\footnote{It is a common sense that intelligent creatures are able to learn concepts from completely unfamiliar environment.}, therefore it has been so far the only known family of implementable harvesting functions\footnote{$V_C$ is a set of vectors.} and it cooperate only with mapping relation ``L". This corollary also explains researchers intuitive preference for linear separability.
\end{enumerate}

\subsection{Explanatory Comment}
This section explains in what sense these two laws and three corollaries above shall be required to be understood. For understanding the essence of these laws and corollaries, it is necessary to experience the detail of what information can be gained and what cannot from the viewpoint of being a mapping relation rather than being the creator of a mapping relation.

\begin{enumerate}
\item Firstly, the definition of a dynamic system shall be generalised \footnote{Not restricted in the dynamicity of  time} as follows:
\\
\textit{A state space $D$ (domain $D \subset R^m$), a set of parameters $\Lambda$, and a rule $M$ that specifies how each state will be mapped to another state space $O$ (range $O \subset R^n$) with the changing of parameters. The rule $M$ is a mapping relation whose domain is $D \times \Lambda$ and whose codomain is $D$. Therefore, $M: D \times \Lambda \rightarrow O (D \subset R^m, O \subset R^n)$. Mapping relation $M$ takes two input $M (x,\alpha)$,  where $x \in D \subset R^m$ is the outcome of each observation and $\alpha \in \Lambda$ is a possible configuration of this mapping relation.
}
\\
The mapping relation discussed in this paper is referred as a certain configuration of a dynamic system $M_{\alpha}(x \mid \alpha \in \Lambda)$, the dynamicity related topic is off-discussion in this paper\footnote{The only discussion that related to the dynamic property is law two.}.

\item For mapping relation $M_L:D \rightarrow O, (D \subset R^m, O \subset R^n)$, $m$, $n$ and the range of each dimension are assumed to be very large number, to infinity ideally \footnote{The scenario shown in figure \ref{strategy1} illustrates the reason why the range of each dimension are ideally to be infinity.}. The reason of this assumption is straightforward.\\
It is apparently that a domain $D \subset R^3$ is less likely to contain less information than only 2 of its dimension are recorded and it is also less likely to contain less information than only the integer part of all dimensions are recorded. And the direct consequence of having a domain which contains only a small amount of information is that there will be not enough information to define possibly different concepts. An intuitive example of this scenario would be when a person with a high degree of myopia accidentally loses his glasses. Since the domain is supposed to be big enough to carry large amounts of information, it is reasonable to assume that the dimensionality of the range is high enough to contain enough parallel hyperplanes, and the range of each dimensionality seems not very important at this stage, but it is directly related to the further analysis of mapping relation $L$. It is worth mentioning that shrinking the range of each dimensionality is clearly a strategy that enable a non-learning model which does not satisfy law one behaves like it is able to learn the domain.

\begin{figure}[H]
    \centering
    \includegraphics[width=140mm]{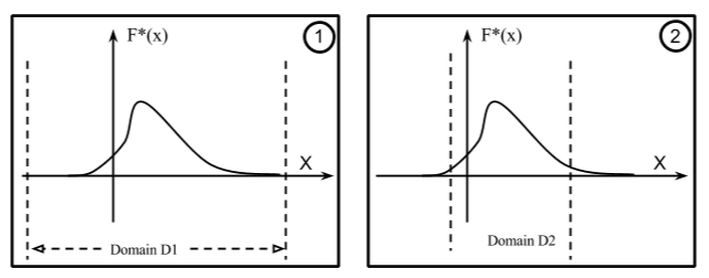}
    \caption{Non-learning model to Learning model}
    \label{strategy1}
\end{figure}

\noindent
As shown in figure \ref{strategy1} (picture 1), the function $F^*$ is defined on domain $D1$, and does not satisfy law one appearantly. But by shrinking the range of the only dimension of its domain, as shown in \ref{strategy1} (picture 2), now $F*$  behaves like it satisfies law one almost perfectly. This strategy is usually being known as normalisation process in the field of machine learning, by doing so, application usually will be fed with pre-prepared data and can behave like it is able to define new concepts based on the appearances of the domain. However this pre-prepared process usually involves extensive knowledge or operation about the input and finial output of the application. Therefore, a learning model is expected to define concepts directly from the observation of high dimensional space with big range(ideally infinite) of each dimension.

\item Law one together with corollary three indicate that harvest function $V_C$ could utilise global representations for all possible concepts without the existence of $L$. This implication seems irrational at first glance, however by assuming the well accepted learning model: human, is learning model as described in corollary three $ V_C \circ L$, the correctness of this implication is undeniable. This implication equivalent to:

\begin{tightcenter}
\textbf{
        \textit{
People can see everything they could possible see in future any time.
                }
        }
\end{tightcenter}

Most people do not realise this fact in their daily life, but this is the foundation of creative activities such as painting or sculpture \footnote{Dynamic property of a system is off-discussion here.}. It is not hard to image that a painter would still be able to create masterpiece once he lose his sight and for normal people who do not master this painting skill the experience of dreaming strange-looking creatures is not unusual. In this case, when $V_C$ and $L$ were being constructed, all dimensions of the domain contain same type of information and the appearances of a concept is fairly straightforward provided. What if the domain consists of different types of information? We know that people who merely remember the answer of certain examination or all past examinations will not be acknowledged as having learned the concept of the corresponding subject, on the other hand people who learned concepts of a certain subject cannot only give answers to every possible related questions, but also can see facts that cannot be seen by people who merely remember those answers. This is an example of non-learning model(memory system) and learning model that are related to high-level concept. In this case a high-level concept will be defined based on different type of information and apparently how different types of information being clustered will largely affect whether higher level concept can be effectively learned or not, and this belongs to the discussion of intelligence which beyond the scope of this paper \footnote{From the viewpoint of a blind person, the description of the painting of ``The Last Supper" is full of articulation problem, equivalently in the teaching-learning process, for student who has no corresponding basic knowledge, the introduction of some concept would be also full of articulation problem. For most people, intelligence is about how different type of information can be related so that higher level concepts could be learned.}.

\item
The generalised definition of Machine Learning Problem is:
\begin{tightcenter}
\textbf{
        \textit{
``Finding desired dependence using a limited number of observations "\cite{vapnik2000nature}
                }
        }
\end{tightcenter}

And this general description consists of three components\cite{vapnik2000nature} as shown in figure \ref{glm}.
\begin{figure}[H]
    \centering
    \includegraphics[width=60mm]{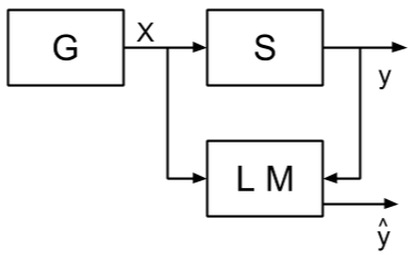}
    \caption{General model of learning from observation\cite{vapnik2000nature}}
    \label{glm}
\end{figure}

\begin{itemize}
\item An generator $(G)$ of vector $x \in R^n$, $x$ is a variable which represents possible outcome of independent observations and presumably $x$ follows an unknown distribution $F(X)$.
\item A supervisor $S$ that takes $x$ as input and returns an output value $y$. The mechanism of the supervisor $F(y \mid x)$ is also unknown.
\item A learning machine that manipulates a family of functions $Loss(y,\hat{y} =f(x,\Theta))$.
\end{itemize}
\noindent
All solutions of machine learning problems directly depend on people's hypothesis of $f(x, \Theta)$ \footnote{ $\Theta$ is the parameter space of function f.}. Based on current observation and by implementing some training algorithm, the goal of training process is to find a desirable $\theta$ so that when facing new observation results the difference between $\hat{y}$ and $y$ can be minimised.

\bigskip
\noindent
In this paper, a mapping relation is referred as a possible configuration of a dynamic system:$M_{\alpha}(x \mid \alpha \in \Lambda)$, therefore if we treat the hypothesis $f(x,\Theta)$ as a dynamic system, then any configuration of $f$: $f_{\theta}(x \mid \theta \in \Theta)$ is within the discussion of this paper. And from the viewpoint of a certain configuration $f_{\theta}(x \mid \theta \in \Theta)$,  there is no difference among various machine learning concepts such as on-line, off-line, supervised, non-supervised, semi-supervised and so no. So law \textit{\rom{1}}, \textit{\rom{2}} and the following corollaries are eligible to be applied on the hypothesis $f(x,\Theta)$ without regarding how a desirable result shall be obtained.

\bigskip
\noindent
The necessary conditions for $f(x,\Theta)$ to be a learning model are law \textit{\rom{1}} and law \textit{\rom{2}}, therefore our hypothesis about observation is the most critical part which affects whether $f(x,\Theta)$ could be a learning model or not. When applying a probabilistic model to solving the object recognition problem, the hypothesis is based on our belief that there are unknown statistical laws which represent different concepts. Assume mapping relation $F_{hypo}$ is able to mimic the unknown statistical law of the appearance of a cup perfectly, then the mapping relation we get is as follows:
\begin{figure}[H]
    \centering
    \includegraphics[width=80mm]{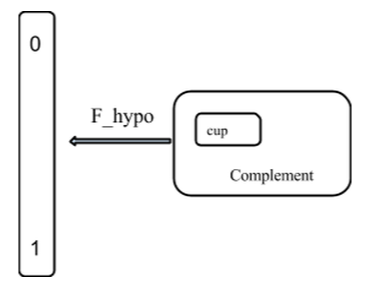}
    \caption{Probability Model: A Cup}
    \label{hypo1}
\end{figure}
\noindent
At this stage, $F_{hypo}(Cup \mid observation ; \theta \in \Theta)$ is able to give the probability of being a cup of each observation. What if we want $F_{hypo}$ to give probabilities of being a dog and a cup of each observation respectively? Being different from our hypothesis in regression problem, there is no real number naturally related to different objects. Therefore we usually choose another way of constructing our hypothesis of observation as shown in figure \ref{hypo2}.

\begin{figure}[H]
    \centering
    \includegraphics[width=100mm]{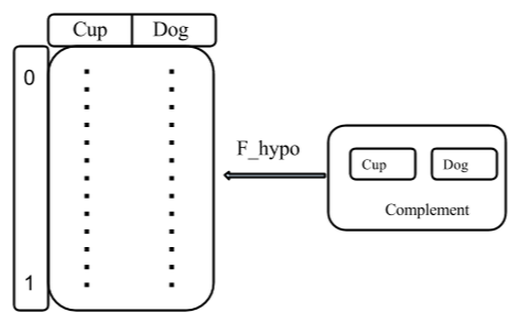}
    \caption{Probability Model: A Cup \& A Dog}
    \label{hypo2}
\end{figure}
\begin{tightcenter}
        \textit{
$F_{hypo}: R^n \rightarrow R \times I$ ($I$ is the indicator set)
                }
\end{tightcenter}
\noindent
However the introduction of this “Indicator Set” breaks law \textit{\rom{2}} directly. The size
of this indicator set  is the first order global information that represents the information of the amount of objects which $F_{hypo}$ can effectively identify, this information is given by us (the creator of this model) and without this information the model cannot reach its best performance after the training process, and the worst case is that the training process will not converge forever\cite{vapnik2000nature}. This example indicates that people should careful construct our hypothesis for getting a learning model, once too much information about the dataset was adopted, the corresponding training process is nothing but a high performance database(non-learning model) constructing program.

\item Although hypothesis $f(x,\Theta)$ being used for  classification problems may not satisfy law \textit{\rom{2}} and \textit{\rom{1}}, the corresponding loss function $Loss(y,f(x,\Theta))$\cite{vapnik2000nature} are almost always learning models: $Loss:X \subset D \times \theta \rightarrow R$. The term ``Learning" in machine learning problem means learning a desirable $\theta \in \Theta$. The term ``Learning" in this paper means the ability of a mapping relation. Furthermore, in the field of classical machine learning, by manipulating $Loss(\Theta \mid X,Y)$ based on a set of observation $X,Y$, our goal is to minimise the loss $r \in R$ during the training process. Hopefully after training, the loss $Loss(y,f(D \setminus X \mid \theta))$ will still be very small. It is obvious that knowledge about $D$ (first order global information) is the key factor, because it is necessary for our hypothesis to have equivalent performance on current observation $X$ and possible future observation $D \setminus X$.

\item Since the existence of concepts are defined based on different appearances, the notion "right" or "wrong" is redundant in this situation. More precisely, for a learning model $M_L:D \rightarrow O$ talking about whether the mapping relation $ x \rightarrow y (x \in X \subset D; y \in Y = M(x))$ is correct or not is meaningless, in other words, any criterion which can be used to test this mapping relation only provides first order global information with respect to $M_L$. For example, When explaining the object recognition problem using the theoretical framework proposed in this paper, two seemingly counter-intuitive deductions are dataset separation and dataset merge problems.
\begin{itemize}
\item Dataset separation problem: When dataset generated by one object is separated, there could be two different sets of invariant representations.
\item Dataset merge problem: When two datasets of different objects are merged together, there could be a new set of invariant representations.
\end{itemize}
\begin{figure}[H]
    \centering
    \includegraphics[width=120mm]{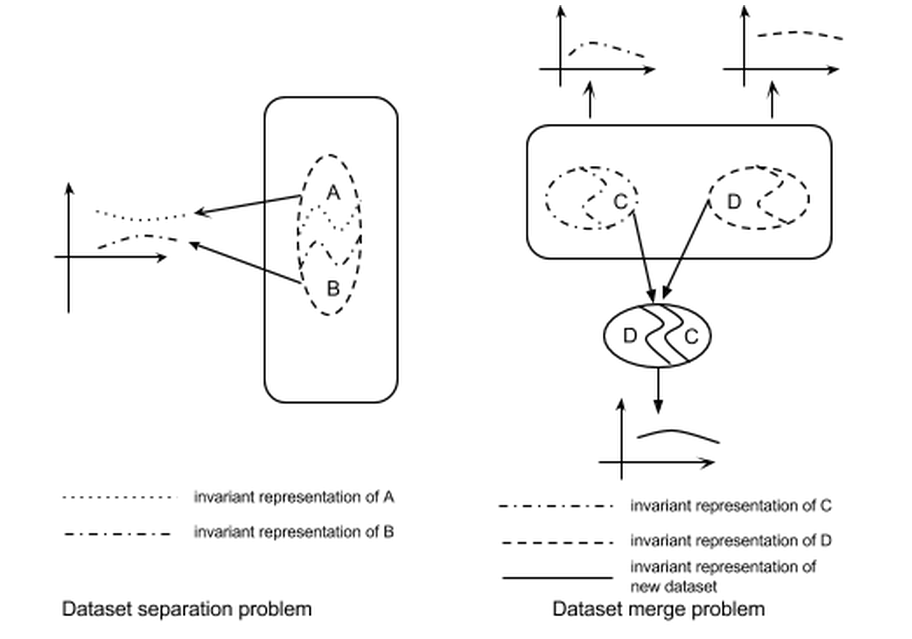}
    \caption{Dataset seperate and merge problems}
    \label{overfitting}
\end{figure}
Above description is the common intuitive understanding of these two problems, but there are two mistakes about this understanding:
\begin{itemize}
\item The information generators is defined by the mapping relation, so these two problems are equivalent.
\item Since there is no observation independent concept, 
solely talking about the existence of certain concept is meaningless and this example shown in figure \ref{overfitting} is often known as over-fitting problem in the field of machine learning. However, from the viewpoint of a learning model there is no notion of ``right" or ``wrong", therefore it is not a problem, it is a phenomenon which can be used to verify the validity of future implementation based on this theoretical framework.
\end{itemize}
\end{enumerate}

\noindent
Up to this stage it has been considered to give a summary which explains relations among each sections of this paper, so that readers could have a better understanding of this paper.

\bigskip
\noindent
Firstly, by giving definitions of local and global information (representations), the intuitive understanding about the relation between observation and the reality can be explained in the framework of mapping relation. Based on this formalised expression of information, law \textit{\rom{1}} indicates that for a mapping relation to be a learning model it at least should not be a information black-hole. Law \textit{\rom{2}} confines possible information being used in constructing the mapping relation so that from a mapping relations point of view it will not be able to utilise information which goes beyond its field of vision (first order global information).Corollary \textit{\rom{1}} proves the existence of a set of equivalent constraints which can harvest global information from a learning model and all possible sets of equivalent constraints are functions. Corollary \textit{\rom{2}} proves a learning model $M_L$, together with any of its possible harvest function $H$ still satisfies law \textit{\rom{1}} and different concepts of any type of information $M_L$ can be represented by different real numbers although there are infinite many possible choices and it could be impossible to find any of them. Corollary \textit{\rom{3}} proves it is possible to have a harvest function $V_C$ which is independent from any possible first order global information and all learning model $M_L$ has a corresponding family of mapping relation $L$ that are also learning models.

\section{Conclusion and Future Works}
\bigskip
\noindent
Focusing on achieving functional similarities has been the basic guideline of top-down approaches since always and various machine learning techniques has been developed under this guideline. However, without a fundamental interpretation about what a given ability is, there could be dramatic differences between constructing a near-this-ability-performance system and constructing a having-this-ability-performance system. It is just like researchers might eventually build a sky-facing Columbiad space gun which could shot three people to the Moon or even the Mars, but this should not bring researchers any confidence to claim that the space gun is the promising technique for interstellar travel. This paper gives two laws which are necessary conditions for any mapping relations to be acknowledged to have learning ability (to be a learning model) and illustrate these following facts:
\begin{itemize}
\item ``Learning" is the ability of identifying the existence of new concept.
\item A mapping relation $M_L$ will be acknowledged as a learning model of information $M_L$ only when it is able to possess this ``learning" ability with no help from other learning models.
\item If the mathematical expression (model) of our hypothesis of observation is not being constructed carefully, information provided by us (the creator of the model) can exceed the vision of the model very easily and causes the model to be a non-learning model, therefore further development based on this model for achieving the human-level-learning ability can lead to only inevitable failure. For example, using labels to represent our observation in typical classification problems will directly leads our hypothesises of observation to failure of satisfying law \textit{\rom{2}} and \textit{\rom{1}}.
\item There exists a common learning model that utilises the power of hyperplanes.
\end{itemize}
By inspecting from the viewpoint of mapping relations and treating them equally (human, animal, apparatus), these key ideas of this paper can be appreciated more clearly.

\bigskip
\noindent
Because the discussion of this paper is confined in certain configurations $\theta$ of a dynamic system $f(x,\Theta)$, so there are two major future works: What is the family of $L(x, \Theta)$; how to get a certain configuration of common learning model $L(\theta \in \Theta)$ and how these parameters change dynamically. Further research on these two issues will address following questions:
\begin{itemize}
\item Under what circumstances is hierarchy structure\footnote{Currently, there is no explanation about the necessity of having layer structure for both biological and artificial neural networks.} of a learning model necessary?
\item In a typical machine learning problem, the learning result is a $\theta \in \Theta$, however, for a learning model, compared with obtaining desirable $\theta$, a more valuable question is ``how a learning model encoding the unknown constraint of local representations of all concepts in the domain?".
\item How could possible existence of concepts depend on the information being contained in the domain quantitatively?
\item The dependency relation between learning model and non-learning model (memory system).
\end{itemize}

\noindent
These days, swing into the saddle does not mean people are going on a long journey; juggling on the pavement is usually just for exercise. After the industrial revolution, people keep inventing all different kinds of machines which enable us to exceed the physical limitation of our biological blueprint. Therefore, the ability of invention, or more broadly speaking, intelligence is the proudest property of human and it has not been simulated by any man-made-machine successfully, yet. This dissertation is one of many steps to the inevitable future when human might not be the absolutely necessary information resource of automatic systems and we could harvest knowledge which cannot be provided by our learning ability.
\newpage
\bibliographystyle{plain}
\bibliography{references}
\end{document}